\documentclass[runningheads]{llncs}
\usepackage[T1]{fontenc}
\usepackage{graphicx,verbatim}
\usepackage{multirow}
\usepackage{hyperref}

\usepackage{dirtytalk}
\usepackage{caption}
\usepackage{subcaption}
\usepackage{arydshln}
\usepackage{array}
\usepackage{svg}
\usepackage{pdfpages}
\usepackage{footmisc}
\usepackage{svg}
\usepackage{mathtools}
\usepackage{amssymb}  
\usepackage{siunitx}
\usepackage{tabularx} 
\usepackage{longtable} 
\usepackage{xltabular}
\usepackage{booktabs}  
\usepackage{color, colortbl}
\usepackage{hhline}
\usepackage[english]{babel}
\usepackage[autostyle]{csquotes}
\usepackage{microtype}
\usepackage{pifont}
\usepackage{cleveref}
\crefname{figure}{Fig.}{Figs.}
\Crefname{figure}{Figure}{Figures}
\crefname{table}{Tab.}{Tabs.}
\Crefname{table}{Table}{Tables}

\usepackage[dvipsnames]{xcolor}
\definecolor{color_vnca}{RGB}{255, 193, 7}
\definecolor{color_vae}{RGB}{26, 83, 255}
\definecolor{color_styleganca}{RGB}{239, 59, 44}
\definecolor{color_cstyleganca}{RGB}{153, 0, 13}

\begin{document}
\title{Coarse to Fine: Iterative Adversarial Neural Cellular Automata for Medical Image Synthesis}
\titlerunning{StyleGANCA}

\author{Anh Thi Luu\inst{1} \and Nick Lemke\inst{1,2} \and Anirban Mukhopadhyay\inst{1}}  
\authorrunning{A. Luu et al.}
\institute{Technical University of Darmstadt, Darmstadt, Germany \\
    \email{anhthi.luu@stud.tu-darmstadt.de} 
\and ImFusion GmbH, Munich, Germany}
  
\maketitle              
\begin{abstract}

Large-scale, publicly available datasets have driven advances in deep learning, but privacy and legal restrictions often limit data sharing in medical imaging.
Synthetic data generation offers a privacy-friendly alternative to enable the training of high-performance models on health data.
While most state-of-the-art generative models produce high-quality images, they remain computationally expensive, which limits their applicability on resource-constrained hardware.
We propose StyleGANCA, the first lightweight general-purpose NCA-based generative adversarial network.
The  architecture integrates a StyleGAN-inspired mapping network and adaptive style modulation into a multi-scale NCA synthesis process, enabling latent-controlled image generation through iterative local interactions. 
We evaluate StyleGANCA on BloodMNIST and PathMNIST against adversarial, variational, diffusion, and NCA-based baselines. 
Experimental results demonstrate that StyleGANCA achieves competitive image quality with substantially fewer parameters than baseline architectures, achieving the best FID and KID scores on PathMNIST with only 617k parameters.
Furthermore, downstream experiments show that the generated images preserve class-specific information and effectively support the training of multi-class classifiers. 
Our code is publicly available at: \url{https://github.com/MECLabTUDA/StyleGANCA}

\end{abstract}

\section{Introduction}

\begin{figure}
    \centering
    \includegraphics[width=\linewidth]{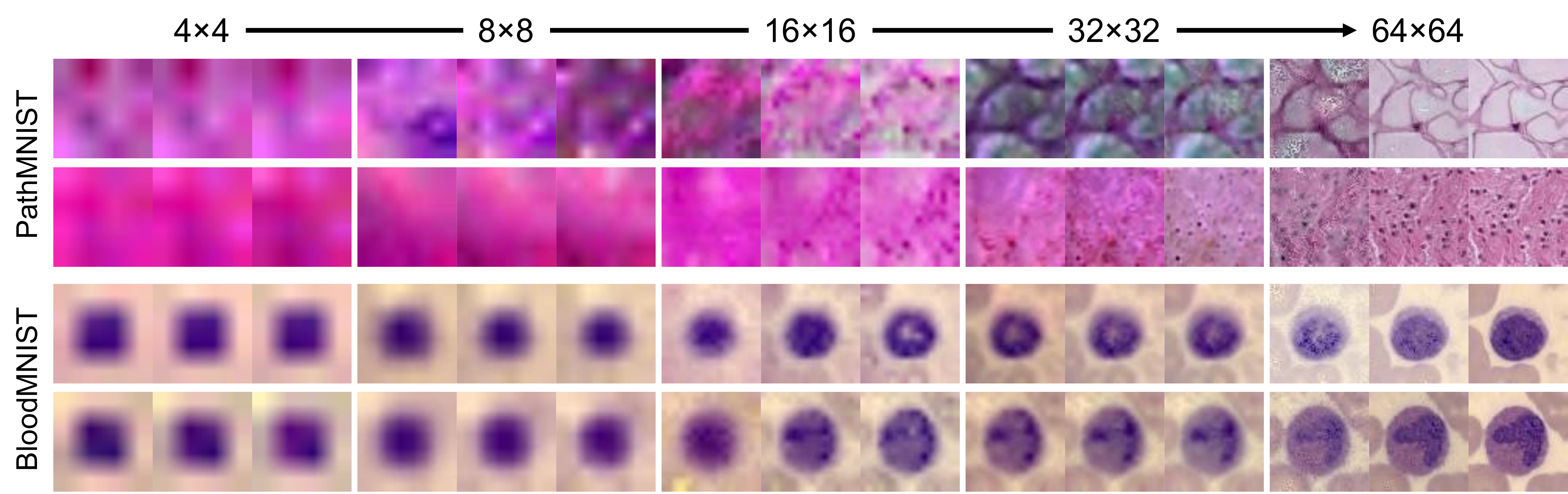}
    \caption{StyleGANCA sample progression across resolutions and update steps.}
    \label{fig:iterative_generation}
\end{figure}

The success of modern deep learning methods largely depends on the availability of large and diverse datasets. 
In medical imaging, access to large-scale datasets is particularly important for training robust and generalizable models across different patient populations and disease characteristics. 
However, collecting and sharing medical imaging data across institutions remains challenging because of legal, ethical, and privacy regulations~\cite{lemke2025equitable,gdpr}.
As a result, synthetic data generation has emerged as a promising approach for enabling model development while reducing dependence on sensitive patient information~\cite{chen2021synthetic}.


While Variational Autoencoders (VAEs) \cite{DBLP:journals/corr/KingmaW13} provide stable training but often struggle to match the perceptual image quality of adversarial approaches, diffusion models \cite{DBLP:conf/nips/HoJA20} achieve high-quality image synthesis at the cost of slow sequences of iterative refinement. 
In contrast, Generative Adversarial Networks (GANs) \cite{DBLP:conf/nips/GoodfellowPMXWOCB14} produce high-quality images at a moderate computational cost. 
Over the past decade, GAN architectures have improved generation quality through deep convolutional architectures \cite{DBLP:conf/iclr/BrockDS19,DBLP:conf/cvpr/KarrasLA19,DBLP:conf/cvpr/KarrasLAHLA20}.
However, these improvements typically require increasingly large models with millions of parameters, motivating the development of more parameter-efficient generative architectures.

Neural Cellular Automata (NCAs) \cite{DBLP:journals/corr/abs-1809-02942,mordvintsev2020growing} provide an alternative generative paradigm based on iterative local interactions. 
Instead of synthesizing an image in a single forward pass, NCAs progressively construct global structures through repetition of a shared local update rule. 
This results in lightweight and parameter-efficient models with strong scalability properties~\cite{DBLP:conf/ipmi/KalkhofGM23}.
However, existing NCA-based generative approaches either suffer from low-quality generation \cite{DBLP:journals/corr/abs-2010-04949,DBLP:conf/iclr/PalmDSR22}, poor generalization \cite{DBLP:journals/corr/abs-2108-04328}, or slow diffusion sequences \cite{kalkhof-fourierdiffnca}.
Moreover, adversarial NCA models have so far been limited to specialized conditional generation tasks~\cite{DBLP:journals/corr/abs-2108-04328}, leaving a general-purpose NCA-based GAN for image synthesis unexplored.

In this work, we introduce \textbf{StyleGANCA}, the first general-purpose multi-scale NCA-based GAN.
The proposed architecture combines StyleGAN-inspired latent control~\cite{DBLP:conf/cvpr/KarrasLA19,DBLP:conf/iccv/HuangB17} with iterative image synthesis through hierarchical NCAs~\cite{lemke2025octreenca}. 
Instead of using latent noise directly, a mapping network transforms latent noise into an intermediate latent representation, which is used to control the features at each synthesis block of the multi-scale NCA synthesis network through adaptive group normalization. 
By assigning dedicated NCA modules to different resolution levels, the model efficiently captures both global image structure and fine-grained details while reducing the parameter count by over 97\% compared to StyleGAN.
The progressive generation scheme of StyleGANCA is visualized in \cref{fig:iterative_generation}.
Furthermore, we extend the framework to a class-conditional variant we term cStyleGANCA.
We evaluate StyleGANCA on PathMNIST and BloodMNIST and compare it against diverse baselines to qualitatively and quantitatively assess its generative capabilities, evaluate the utility of the generated images for downstream classification, and understand the impact of the chosen hyperparameters.
Experimental results demonstrate competitive image quality and strong parameter efficiency. 

\section{Methodology}
\begin{figure}[t]
    \centering
    \caption{Overview of the proposed StyleGANCA generator.}
    \includegraphics[width=1\linewidth]{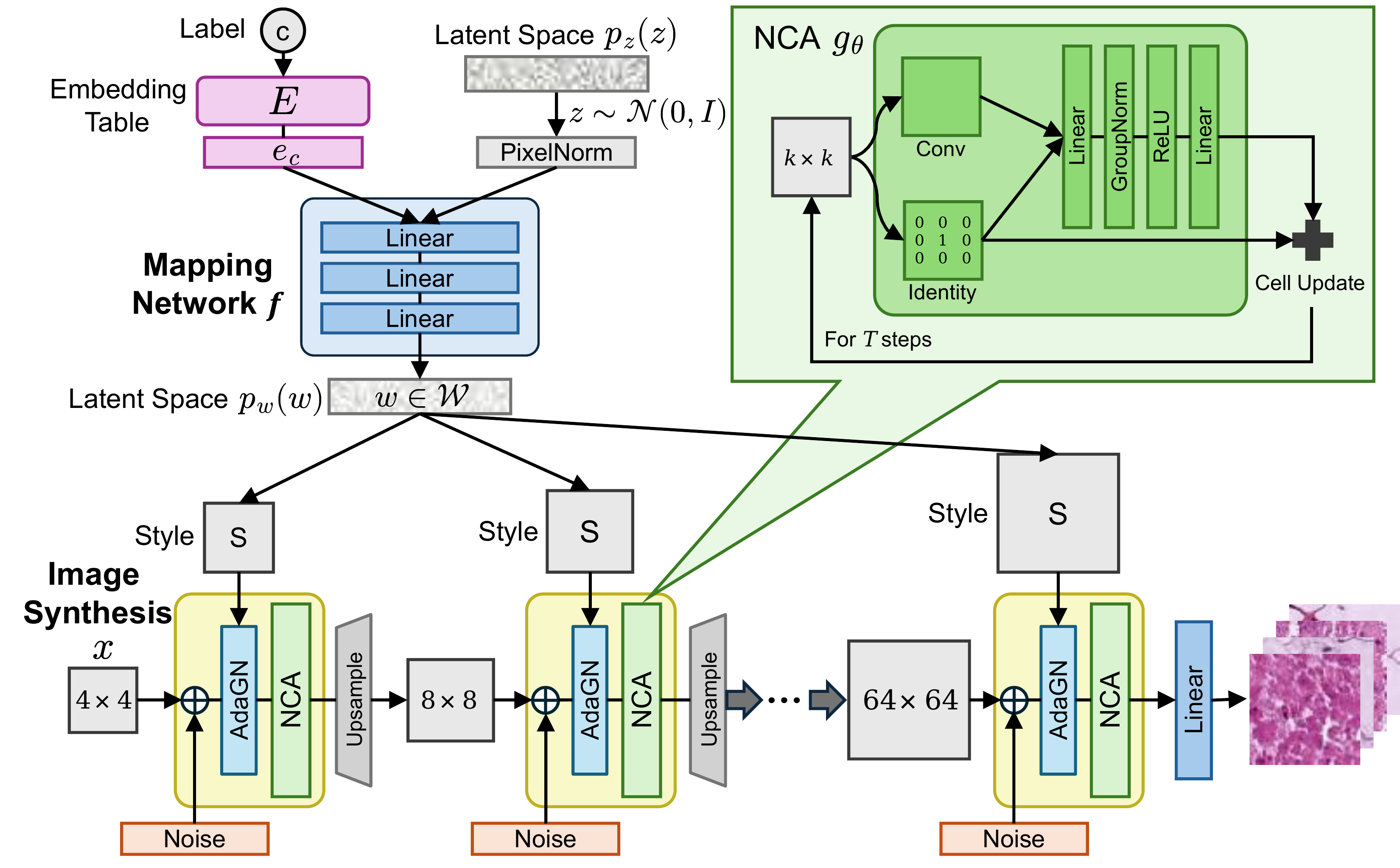}
    \label{fig:styleganca}
\end{figure}

Traditional GANs \cite{DBLP:conf/cvpr/KarrasLA19,DBLP:journals/corr/RadfordMC15,DBLP:conf/icml/ArjovskyCB17} synthesize images through a sequence of feed-forward convolutional layers.
We depart from this idea by replacing the synthesis network with a hierarchy of NCAs operating across multiple spatial resolutions.
The generator $G$ consists of two main components: a latent mapping network and a multi-scale NCA-based synthesis network.
The resulting architecture, StyleGANCA, combines style-based latent control with iterative image generation through local interactions.
An overview of the proposed architecture is shown in \cref{fig:styleganca}. 
The generator $G$ is trained adversarially against a convolutional discriminator $D$ that distinguishes generated images from real images.
Specifically, we employ the discriminator architecture proposed in DCGAN~\cite{DBLP:journals/corr/RadfordMC15}, consisting of 5 convolutional layers.

\subsection{Latent Mapping Network}
Following StyleGAN \cite{DBLP:conf/cvpr/KarrasLA19}, the input latent vector $z \sim \mathcal{N}(0,I)$ is first transformed into an intermediate latent representation $w = f(z) \in \mathcal{W}$, where $f$ denotes a non-linear fully connected mapping network.
The mapping network $f$ disentangles high-level attributes to control the style information used throughout the image synthesis.
Rather than directly influencing the generator input, the latent code is used to produce style parameters that control the synthesis networks at every resolution level.

\subsection{Multi-Scale NCA-based Synthesis Network}
Rather than generating images directly from the latent vector, the synthesis process starts from a learned constant state $x \in \mathbb{R}^{4\times4\times C}$, an approach inspired by StyleGAN, where $C$ denotes the number of hidden channels.
The synthesis network consists of a hierarchy of synthesis blocks, each operating at progressively increasing resolution levels.
Each block refines the current NCA state before it is upsampled and propagated to the next block. 
Starting at low spatial resolution allows global structures to emerge, while higher-resolution blocks refine local textural fine-grained details.
The final image representation is then mapped to the image space using a fully connected output layer.
Each synthesis block operates on a state representation and consists of three operations: noise injection, style modulation, and NCA state updates.

\textbf{Noise Injection} introduces stochastic variation over fine-grained details during image generation to improve image diversity.
For each synthesis block, Gaussian noise $n \sim \mathcal{N}(0,I)$ is sampled and added to the hidden state using learnable per-channel scaling factors, resulting in the perturbed state
$ \tilde{x} = x + \alpha \cdot n, $
where $x$ denotes the current hidden state and $\alpha$ controls the influence of the noise. 

\textbf{Style Modulation} allows the latent representation to influence the evolution of the image.
The perturbed state maps $\tilde{x}$ are modulated using style information obtained by transforming the intermediate latent representation $w \in \mathcal{W}$ into scale and bias parameters $s_{\text{scale}}, s_{\text{bias}}$: 
\begin{align}
    s_{\text{scale}} = f_{\text{scale}}(w), && s_{\text{bias}} = f_{\text{bias}}(w), && \text{with } f_{\text{scale}},f_{\text{bias}}: \mathbb{R}^{d_w} \rightarrow \mathbb{R}^C,
\end{align}
where ${d_w}$ denotes the dimensionality of $w \in \mathcal{W}$.
The hidden state is first normalized using Group Normalization~\cite{DBLP:conf/eccv/WuH18} and then modulated through Adaptive Group Normalization (AdaGN):
\begin{align}
    \text{AdaGN}(\hat{x}, (s_{\text{scale}}, s_{\text{bias}})) = s_{\text{scale}} \cdot \hat{x} + s_{\text{bias}},
\end{align}
where $\hat{x}$ denotes the normalized state representation.
Unlike StyleGAN, which employs Adaptive Instance Normalization, we adopt Adaptive Group Normalization to preserve correlations between channels during the iterative NCA updates.

\textbf{NCA Updates} propagate information across the AdaGN-modulated state map through repeated local interactions and gradually refine the image.
At each step $t$ over a fixed number of steps $T$, neighborhood information is first extracted from the current hidden state $x_t$ using a $3\times3$ convolution $\mathrm{N}(x_t)$.
The resulting neighborhood representation together with the hidden state is processed by a lightweight neural network $g_{\theta}$ to compute a state increment.
The hidden state is then updated according to
\begin{align}
    x_{t+1}=x_t + g_{\theta}(x_t, \mathrm{N}(x_t)), 
\end{align}
where $g_{\theta}$ represents the shared neural update function.
Each resolution level learns an independent NCA with its own parameters.
Therefore, lower-resolution NCAs primarily capture coarse image structure and long-range dependencies, whereas higher-resolution NCAs focus on textures and fine-grained details.
The implementation of the NCA update function can be seen in \cref{fig:styleganca}.

\subsection{Conditional Generation}
We extend our class-agnostic StyleGANCA by a conditional variant.
The conditioning label $c$ is mapped to a learnable embedding and concatenated with the latent vector $z$ before the mapping network $f$.
The resulting conditional intermediate latent representation is then used to control the synthesis process at all resolution levels.
Injecting conditional information before the mapping network allows class-specific information to influence image generation from the earliest to the latest stage of synthesis.

The discriminator receives the same conditional information through a learned label embedding that is combined with the image representation before prediction, such that the discriminator evaluates the image in the context of the label. 
Apart from the incorporation of label embeddings, the architecture and training procedure remain identical to the unconditional setting.
\section{Experimental Setup}
In this section, we describe our experimental setup, including the training datasets, training settings, baseline generators, and evaluation metrics.

\textbf{Datasets:} 
We evaluate StyleGANCA on the two medical image datasets BloodMNIST~\cite{bloodmnist.2020.acevedo} and PathMNIST~\cite{pathmnist.2018.kather,pathmnist.2019.kather}.
BloodMNIST~\cite{bloodmnist.2020.acevedo} consists of microscopic blood cell images from eight classes. 
PathMNIST~\cite{pathmnist.2018.kather,pathmnist.2019.kather} contains colorectal histopathology image patches from nine tissue classes. 
For generator training, we use the provided images at a resolution of $64 \times 64$ pixels. 
For the downstream multi-class classification experiments, classifiers are trained and evaluated on the $28 \times 28$ images, following the evaluation protocol of the MedMNIST benchmark.
We use the predefined data splits from MedMNNIST.

\textbf{Training Details:}
StyleGANCA is optimized using the Wasserstein GAN objective \cite{DBLP:conf/icml/ArjovskyCB17} with gradient penalty (WGAN-GP) \cite{DBLP:conf/nips/GulrajaniAADC17}. 
Training is performed using the Adam optimizer with learning rates of $1\times10^{-4}$ for the generator and $2\times10^{-4}$ for the discriminator and Adam parameters $\beta_1 = 0.0$ and $\beta_2 = 0.9$, following the WGAN-GP training procedure.
Batch size is set to 32 for all experiments.
The generator employs an exponential moving average (EMA) of the model parameters for evaluation. 
To ensure training convergence, StyleGANCA is trained for 200 epochs.
StyleGANCA uses five hierarchical synthesis blocks.
Other hyperparameters are found in the ablation study.
The discriminator uses a base feature size of 64 and is updated five times for every generator update with a gradient penalty coefficient of $\lambda = 10$,  ensuring a reliable estimate of the Wasserstein distance~\cite{DBLP:conf/icml/ArjovskyCB17,DBLP:conf/nips/GulrajaniAADC17}.

\textbf{Evaluation Metrics:}
Image quality is evaluated using Fréchet Inception Distance (FID)~\cite{DBLP:conf/nips/HeuselRUNH17} and Kernel Inception Distance (KID)~\cite{DBLP:conf/iclr/BinkowskiSAG18}. 
Both metrics measure the similarity between the feature distributions of real and generated images based on the Inceptionv3~\cite{DBLP:conf/cvpr/SzegedyLJSRAEVR15} model.
We compare 2048 generated images against an equal number of real images from the test split.
For the multi-class classification experiments, we report the Area Under the ROC Curve (AUC), the accuracy (ACC), and the F1 score.
Note that for the unconditional generative baselines, we pseudo-label the generated samples using a classifier trained on the corresponding real training set. 

\textbf{Baselines:}
We compare StyleGANCA against a VAE~\cite{DBLP:journals/corr/KingmaW13}, UNet denoising diffusion probabilistic model (DDPM)~\cite{DBLP:conf/nips/HoJA20}, Diffusion Transformer (DiT)~\cite{peebles2023scalable},  DCGAN~\cite{DBLP:journals/corr/RadfordMC15}, StyleGAN~\cite{DBLP:conf/cvpr/KarrasLA19}, as well as, NCA-based generative models, such as VNCA~\cite{DBLP:conf/iclr/PalmDSR22} and FourierDiff-NCA~\cite{kalkhof-fourierdiffnca}.

\section{Results}

This section summarizes the results of our experiments, providing a quantitative and qualitative comparison, results on a downstream task and an ablation study.

\begin{figure}[t]
    \centering
    \includegraphics[width=\linewidth]{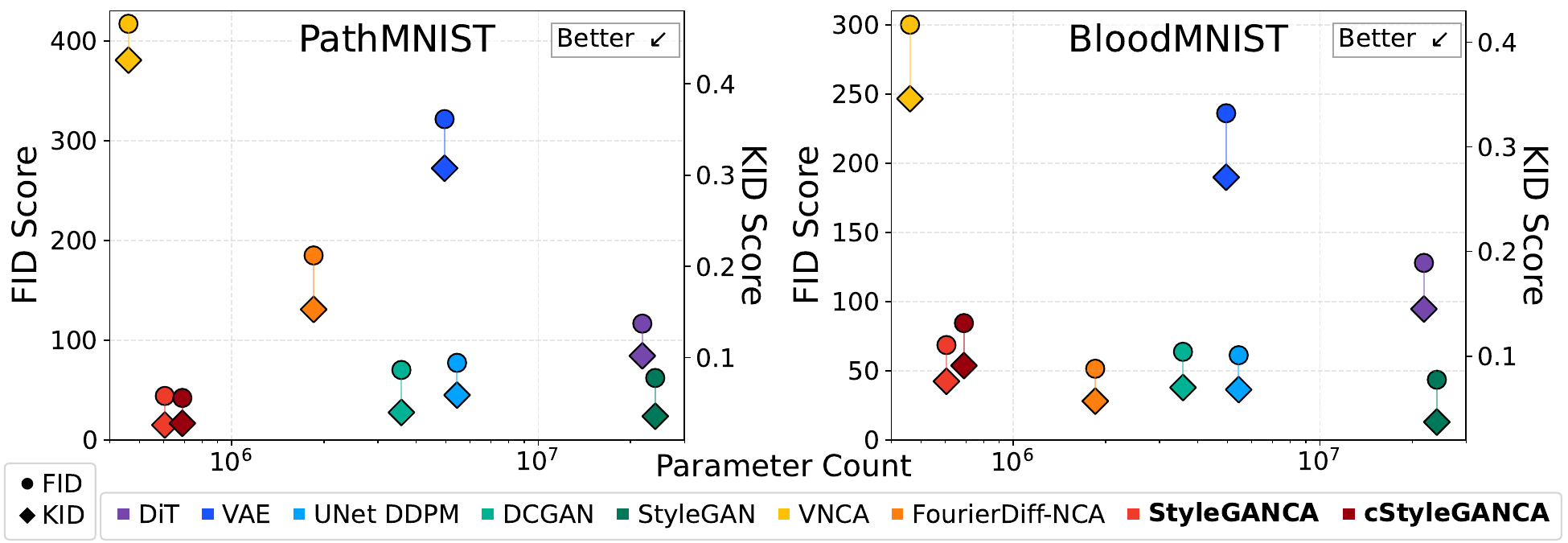}
    \caption{Comparison of image quality and parameter efficiency on PathMNIST and BloodMNIST. FID (circles, left axis) and KID (diamonds, right axis) are plotted against the number of trainable parameters.}
    \label{fig:quantitative-comparison}
\end{figure}

\textbf{Quantitative Comparison:}
\Cref{fig:quantitative-comparison} compares image quality and parameter efficiency on PathMNIST and BloodMNIST. 
On PathMNIST, StyleGANCA~(\textcolor{color_styleganca}{red}) achieves the best overall performance among all evaluated methods.
The conditional variant (\textcolor{color_cstyleganca}{brown}) further improves performance, achieving the lowest FID score while maintaining a similarly low KID score.
While some baselines achieve slightly better image quality on BloodMNIST, StyleGANCA outperforms the VAE (\textcolor{color_vae}{blue}) and VNCA (\textcolor{color_vnca}{yellow}) and performs in a comparable region to the other diffusion- and GAN-based approaches with only 617k parameters. 
Overall, the results suggest that StyleGANCA provides a good trade-off between image quality and model complexity, achieving competitive performance on BloodMNIST and state-of-the-art performance on PathMNIST despite its compact architecture.

\begin{figure}[t]
    \centering
    \includegraphics[width=\linewidth]{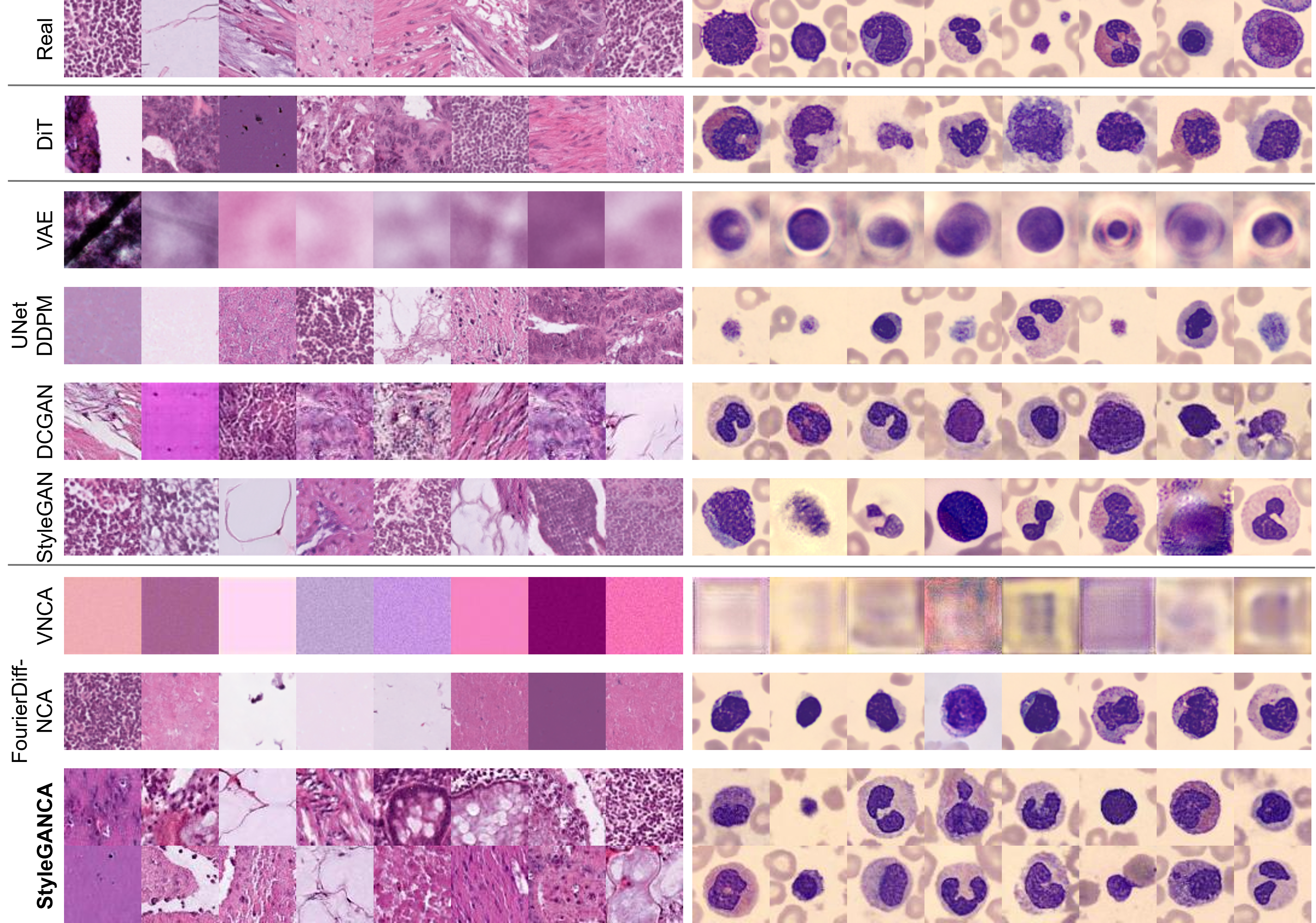}
    \caption{Generated images on PathMNIST (left) and BloodMNIST (right).}
    \label{fig:qualitative-res}
\end{figure}

\textbf{Qualitative Comparison:}
\Cref{fig:qualitative-res} presents samples generated by all models for both datasets. 
Except for VAE and VNCA, all models produce realistic results with only minor perceptual differences from the real samples.
StyleGANCA generates images that are visually comparable to those of significantly larger generative models while maintaining the parameter efficiency of NCAs.
The progressive synthesis process shown in \cref{fig:iterative_generation} demonstrates how the proposed multi-scale architecture first establishes coarse image structures before iteratively refining fine-grained textures, resulting in the final image quality observed in \cref{fig:qualitative-res}.

\begin{table}[t]
    \centering
    \caption{Downstream multi-class classification performance after training on synthetic data. *means data was pseudo-labeled.}

\begin{tabular}{|c|c|c|ccc|ccc|}
\hline
\multirow{2}{*}{\begin{tabular}[c]{@{}c@{}}Model\\ Family\end{tabular}} &                  &  Num.            & \multicolumn{3}{c|}{PathMNIST}                   & \multicolumn{3}{c|}{BloodMNIST}                  \\
                                                                        & Training Data    & Params           & AUC            & ACC            & F1             & AUC            & ACC            & F1             \\ \hline
-                                                                       & Real             & -                & 0.983          & \textbf{0.907} & \textbf{0.863} & \textbf{0.998} & \textbf{0.958} & \textbf{0.955} \\ \hline
Transf.-bsd                                                             & DiT*             & 21,859,248       & 0.973          & 0.838          & 0.778          & 0.989          & 0.885          & 0.865          \\ \hline
\multirow{4}{*}{Conv.-bsd}                                              & VAE*             & 9,903,588        & 0.742          & 0.279          & 0.171          & 0.976          & 0.808          & 0.798          \\
                                                                        & UNet DDPM*       & 5,434,883        & 0.985          & 0.856          & 0.797          & 0.9656         & 0.811          & 0.771          \\
                                                                        & DCGAN*           & 3,576,704        & 0.940          & 0.652          & 0.532          & 0.987          & 0.868          & 0.851          \\
                                                                        & StyleGAN*        & 24,087,311       & 0.984          & 0.874          & 0.834          & 0.986          & 0.878          & 0.856          \\ \hline
\multirow{4}{*}{NCA-bsd}                                                & VNCA*            & 9,732,416        & 0.720          & 0.157          & 0.113          & 0.947          & 0.653          & 0.534          \\
                                                                        & FourierDiff-NCA* & 1,849,632        & 0.943          & 0.681          & 0.599          & 0.988          & 0.871          & 0.850          \\
                                                                        & \textbf{StyleGANCA*}      & \textbf{616,835} & 0.955          & 0.761          & 0.679          & 0.988          & 0.880          & 0.862          \\
                                                                        & \textbf{cStyleGANCA}      & 633,795          & \textbf{0.989} & 0.866          & 0.829          & 0.968          & 0.767          & 0.737          \\ \hline
\end{tabular}
    \label{tab:downstream}
\end{table}

\textbf{Classification Suitability:}
\Cref{tab:downstream} evaluates the practical utility of the generated images on a downstream multi-class classification task. 
For each generative model, a classifier is trained exclusively on synthetic samples and evaluated on the corresponding real test set.
StyleGANCA achieves AUCs of 0.955 and 0.988, corresponding to accuracies of 0.761 and 0.880 on PathMNIST and BloodMNIST, respectively, outperforming previous NCA-based methods while remaining competitive with larger GAN- and diffusion-based models.
These results demonstrate that the generated images preserve meaningful class-discriminative information suitable for downstream learning.



\textbf{Ablation Study:}
To investigate the influence of key architectural hyperparameters, we perform an ablation study, which can be found in \cref{tab:ablation}.
We modify the latent dimension of the NCA linear update layer, the number of NCA state channels, and the number of NCA update steps. 
Increasing each of these hyperparameters consistently improves the FID and KID scores, although at the cost of increased model complexity or GPU memory consumption.
These results show that StyleGANCA offers a flexible trade-off between computational efficiency and image quality, allowing the architecture to be adapted to different resource constraints.

\begin{table}[t]
\centering
\caption{Ablation study of model hyperparameters. \underline{Underlined} text indicates the standard hyperparameter.}
\begin{subtable}[t]{0.32\linewidth}
    \centering
    \caption{Latent Dimension}
    \resizebox{\linewidth}{!}{%
    \begin{tabular}{|c|c|cc|}
\hline
\multirow{2}{*}{\begin{tabular}[c]{@{}c@{}}Latent\\ Dim.\end{tabular}} &             Num.               & \multicolumn{2}{c|}{Metrics}         \\
                                                                                 & Params                  & FID $\downarrow$ & KID $\downarrow$  \\ \hline
64      & \textbf{309,059}           & 63.34            & 0.074  \\
128     & 387,075                    & 62.89            & 0.071\\
\underline{256}     & 616,835                    & 61.68            & \textbf{0.070} \\
512     & 1,371,267                  & \textbf{61.64}   & \textbf{0.070} \\ \hline
\end{tabular}
    }
    \label{tab:ablation-latent}
\end{subtable}
\begin{subtable}[t]{0.32\linewidth}
    \centering
    \caption{State Channels}
    \resizebox{\linewidth}{!}{%
    \begin{tabular}{|c|c|cc|}
\hline
         &    Num.       & \multicolumn{2}{c|}{Metrics}         \\
Chan. & Params & FID $\downarrow$ & KID $\downarrow$  \\ \hline
16       & \textbf{305,123}   & 86.09           & 0.099 \\
32       & 409,027   & 71.07                    & 0.078 \\
\underline{64}       & 616,835   & 65.68                    & 0.076  \\
128      & 1,032,451 & \textbf{59.38 }          & \textbf{0.071} \\ \hline
\end{tabular}
    }
    \label{tab:ablation-channels}
\end{subtable}
\begin{subtable}[t]{0.32\linewidth}
    \centering
    \caption{Update Steps}
    \resizebox{\linewidth}{!}{%
    \begin{tabular}{|c|c|cc|}
\hline
\multirow{2}{*}{\begin{tabular}[c]{@{}c@{}}Steps\\ $T$\end{tabular}} & \multirow{2}{*}{\begin{tabular}[c]{@{}c@{}}GPU Memory\\ (in MiB)\end{tabular}} & \multicolumn{2}{c|}{Metrics}                              \\
                                                                        &                                                                                & \multicolumn{1}{c}{FID $\downarrow$} & KID $\downarrow$  \\ \hline
1        & \textbf{2,158}   & \multicolumn{1}{c}{208.77}           & 0.329 \\
5        & 4,162            & \multicolumn{1}{c}{90.35}            & 0.104  \\
\underline{10}       & 6,760            & \multicolumn{1}{c}{65.68}            & 0.076  \\
15       & 9,376            & \multicolumn{1}{c}{61.54}            & 0.072  \\
20       & 12,004           & \multicolumn{1}{c}{\textbf{59.85}}   & \textbf{0.068} \\ \hline
\end{tabular}
    }
    \label{tab:ablation-steps}
\end{subtable}
\label{tab:ablation}
\end{table}

\newpage
\section{Conclusion}

In this work, we introduced StyleGANCA, a lightweight generative framework that combines style-based latent control with multi-scale NCAs.
By integrating a StyleGAN-inspired mapping network and adaptive style modulation into a multi-scale NCA synthesis architecture, the proposed approach enables controllable image generation through iterative local interactions with merely 617k parameters.

Experiments on BloodMNIST and PathMNIST demonstrate that StyleGANCA achieves a good trade-off between image quality and model complexity.
The proposed model outperforms the baseline approaches on PathMNIST and achieves competitive performance on BloodMNIST with only 617k trainable parameters, which is less than 3\% of the parameters of StyleGAN.
Qualitative results further show that StyleGANCA generates realistic medical images with coherent structures and fine-grained textures, while downstream classification experiments confirm that the generated images preserve meaningful semantic and class-discriminative information.
These results highlight the potential of StyleGANCA as a promising alternative method for controllable and efficient image synthesis.

%
%
%
%
\bibliographystyle{splncs04}
\bibliography{references}
\end{document}